\documentclass[sigconf]{acmart}
\renewcommand\footnotetextcopyrightpermission[1]{} 
\AtBeginDocument{%
  \providecommand\BibTeX{{%
    \normalfont B\kern-0.5em{\scshape i\kern-0.25em b}\kern-0.8em\TeX}}}

\setcopyright{acmcopyright}
\copyrightyear{2018}
\acmYear{2018}
\acmDOI{XXXXXXX.XXXXXXX}

\acmConference[Conference acronym 'XX]{Make sure to enter the correct
  conference title from your rights confirmation emai}{June 03--05,
  2018}{Woodstock, NY}
\acmPrice{15.00}
\acmISBN{978-1-4503-XXXX-X/18/06}

\usepackage{multirow}
\usepackage{graphicx}  
\usepackage{booktabs}  

\begin{document}

\title{Utilizing Large Language Models for Natural Interface to Pharmacology Databases}

\author{Hong Lu, Chuan Li, Yinheng Li, Jie Zhao}
\email{{honglu, chuanl, yinhengli, zhaojie}@microsoft.com}




\begin{abstract}
The drug development process necessitates that pharmacologists undertake various tasks, such as reviewing literature, formulating hypotheses, designing experiments, and interpreting results. Each stage requires accessing and querying vast amounts of information. In this abstract, we introduce a Large Language Model (LLM)-based Natural Language Interface designed to interact with structured information stored in databases. Our experiments demonstrate the feasibility and effectiveness of the proposed framework. This framework can generalize to query a wide range of pharmaceutical data and knowledge bases.
\end{abstract}

\maketitle

\vspace{-0.2cm}
\section{Introduction}

Numerous databases are commonly used in pharmacology \cite{mo-etal-2022-knowledge, chandak2023building}, encompassing various properties of biological entities (such as genes, proteins, drugs, phenotypes, and diseases) and their relationships. Simple queries might involve retrieving GO terms for a specific protein, while complex queries could necessitate analyzing diseases associated with the dysregulation of a genetic pathway. However, formulating SQL queries and comprehending these databases can be challenging and time-consuming for pharmacologists \cite{pourreza2023dinsql, huang2021knowledge}. 
In this work, we study employing Large Language Models (LLMs) to develop a natural language interface that enables pharmacologists to query public or private databases.

\vspace{-0.2cm}
\section{Method}

\begin{figure}[t]  
    \centering  
    \includegraphics[trim=3cm 7cm 1cm 4.2cm, clip, width=0.6\textwidth]{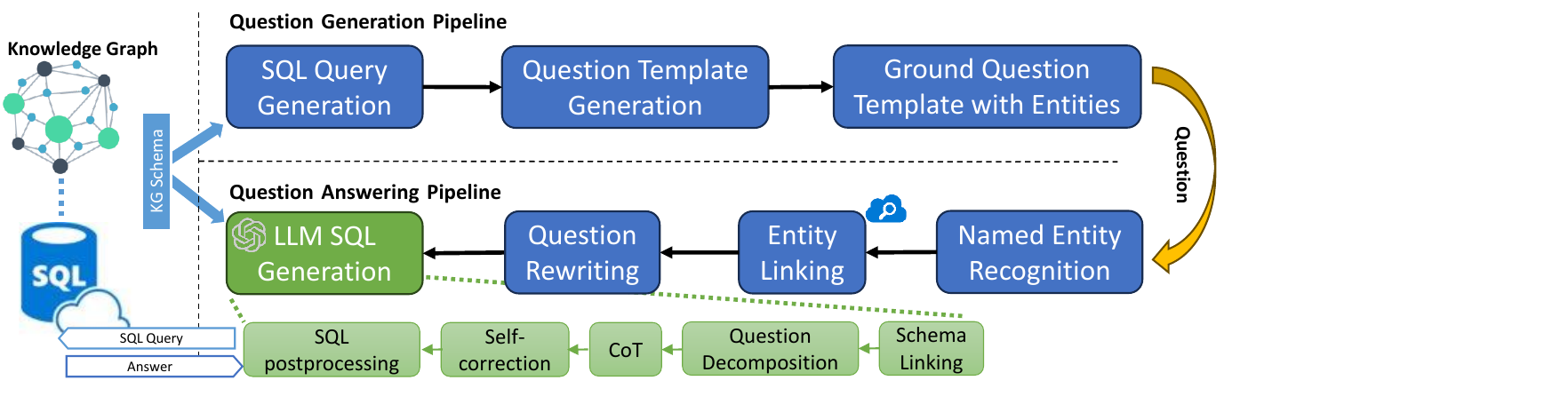}  
    \vspace{-0.9cm}
    \caption{The proposed framework.}  
    \label{fig:pipeline}  
\end{figure}

As illustrated in Fig.~\ref{fig:pipeline}, our end-to-end Question Answering (QA) system allows users to ask natural language questions, which are then translated into SQL queries by GPT-4. Using these queries, the SQL database retrieves relevant information from the databases. Our system employs an in-house Named Entity Recognition (NER) model, followed by an entity linker built on Cognitive Search. Detected entities are replaced with their type-aware placeholders. The GPT-4 powered SQL generator facilitates schema linking, question decomposition, Chain-of-Thought, and self-correction of the generated SQL query.

In addition to the QA system, there is a separate question generation pipeline. This pipeline serves two purposes: generating a synthetic dataset to evaluate the QA model and providing in-context demonstrations for few-shot QA. This dual functionality ensures that the system is effective, accurate, and user-friendly, streamlining the research process and enhancing data exploration efficiency.

\vspace{-0.2cm}
\section{Results}
We evaluated our solution using both synthetic and realistic questions. Our synthetic dataset comprises 60 single-hop questions and 204 two-hop questions, based on the knowledge graph of entities. Domain experts reviewed these generated questions to ensure their quality and relevance. Furthermore, our realistic dataset includes 50 question-answer pairs curated by experts who are unaware of the database schema. This dataset represents the system's effectiveness and applicability to real-world scenarios, further demonstrating its practical utility.

We utilize PrimeKG\cite{chandak2023building} as the knowledge graph, which comprehensively includes 10 types of entities and 30 types of relations with over 8 million edges, aggregated from 20 bio-medical data sources. We employ Exact Match (EM) and F1-score as evaluation metrics. To assess the importance of each module, we conducted an ablation test and evaluated model performance on the realistic dataset with and without named entity recognition (NER) or self-correction (Table\ref{tab:realistic}). This test highlights the significance of both modules. Our model also demonstrated reasonable performance on synthetic questions (Table\ref{tab:synthetic}). The results indicate that NER is a bottleneck and confirm that two-hop reasoning can be more challenging for LLMs.


\begin{table}[h]  
\centering  
\caption{Realistic dataset ablation results. "$-$" denotes removing NER (using oracle entity names) or self-correction (SC).}  
\vspace{-0.4cm}
\label{tab:realistic}
\resizebox{0.7\columnwidth}{!}{
\begin{tabular}{l|l l|l l}  
\toprule  
\multirow{2}{*}{Setting} & \multicolumn{2}{c|}{ChatGPT} & \multicolumn{2}{c}{GPT-4} \\ \cmidrule(lr){2-3} \cmidrule(lr){4-5}  
& EM           & F1           & EM         & F1         \\ \midrule  
Full                         & 0.396            & 0.433            & 0.708          & 0.712        \\  
\quad$-$~{\footnotesize NER} & 0.458            & 0.491            & 0.792          & 0.797          \\  
\quad$-$~{\footnotesize SC}  & 0.313            & 0.316            & 0.708          & 0.712          \\  
\quad$-$~{\footnotesize NER} $-$~{\footnotesize SC} & 0.375            & 0.399            &  0.792         & 0.797          \\ \bottomrule  
\end{tabular}  
}
\end{table} 

\vspace{-0.2cm}
  
\begin{table}[h]    
\centering    
\caption{Evaluation on the synthetic dataset.}    
\vspace{-0.4cm}
\label{tab:synthetic}
\resizebox{\columnwidth}{!}{  
\begin{tabular}{l|l|ll|ll|ll}    
\toprule    
\multirow{2}{*}{LLM} & \multirow{2}{*}{Setting} & \multicolumn{2}{c|}{single-hop} & \multicolumn{2}{c|}{two-hop} & \multicolumn{2}{c}{Overall} \\ \cline{3-8}     
& & EM          & F1          & EM          & F1         & EM          & F1         \\ \midrule    
\multirow{2}{*}{GPT-4}     & Full           & 0.450      & 0.496      & 0.320    & 0.343   & 0.378    & 0.350    \\ \cline{2-8}     
                          & Full $-$~{\footnotesize NER}   & 0.700      & 0.746      & 0.458    & 0.483   & 0.513    & 0.543    \\   
\bottomrule                           
\end{tabular}    
}  
\end{table}

\bibliographystyle{ACM-Reference-Format}
\bibliography{main}


\begin{thebibliography}{4}


\ifx \showCODEN    \undefined \def \showCODEN     #1{\unskip}     \fi
\ifx \showDOI      \undefined \def \showDOI       #1{#1}\fi
\ifx \showISBNx    \undefined \def \showISBNx     #1{\unskip}     \fi
\ifx \showISBNxiii \undefined \def \showISBNxiii  #1{\unskip}     \fi
\ifx \showISSN     \undefined \def \showISSN      #1{\unskip}     \fi
\ifx \showLCCN     \undefined \def \showLCCN      #1{\unskip}     \fi
\ifx \shownote     \undefined \def \shownote      #1{#1}          \fi
\ifx \showarticletitle \undefined \def \showarticletitle #1{#1}   \fi
\ifx \showURL      \undefined \def \showURL       {\relax}        \fi
\providecommand\bibfield[2]{#2}
\providecommand\bibinfo[2]{#2}
\providecommand\natexlab[1]{#1}
\providecommand\showeprint[2][]{arXiv:#2}

\bibitem[Chandak et~al\mbox{.}(2023)]%
        {chandak2023building}
\bibfield{author}{\bibinfo{person}{P. Chandak}, \bibinfo{person}{K. Huang},
  {and} \bibinfo{person}{M. Zitnik}.} \bibinfo{year}{2023}\natexlab{}.
\newblock \showarticletitle{Building a knowledge graph to enable precision
  medicine}.
\newblock \bibinfo{journal}{\emph{Sci Data}} (\bibinfo{year}{2023}).
\newblock


\bibitem[Huang et~al\mbox{.}(2021)]%
        {huang2021knowledge}
\bibfield{author}{\bibinfo{person}{X. Huang}, \bibinfo{person}{J. Zhang},
  \bibinfo{person}{Z. Xu}, \bibinfo{person}{L. Ou}, {and} \bibinfo{person}{J.
  Tong}.} \bibinfo{year}{2021}\natexlab{}.
\newblock \showarticletitle{A knowledge graph based question answering method
  for medical domain}.
\newblock \bibinfo{journal}{\emph{PeerJ Comput Sci}} (\bibinfo{year}{2021}).
\newblock


\bibitem[Mo et~al\mbox{.}(2022)]%
        {mo-etal-2022-knowledge}
\bibfield{author}{\bibinfo{person}{L. Mo}, \bibinfo{person}{Z. Wang},
  \bibinfo{person}{J. Zhao}, {and} \bibinfo{person}{H. Sun}.}
  \bibinfo{year}{2022}\natexlab{}.
\newblock \showarticletitle{Knowledge Transfer between Structured and
  Unstructured Sources for Complex Question Answering}. In
  \bibinfo{booktitle}{\emph{SUKI}}.
\newblock


\bibitem[Pourreza and Rafiei(2023)]%
        {pourreza2023dinsql}
\bibfield{author}{\bibinfo{person}{M. Pourreza} {and} \bibinfo{person}{D.
  Rafiei}.} \bibinfo{year}{2023}\natexlab{}.
\newblock \bibinfo{title}{DIN-SQL: Decomposed In-Context Learning of
  Text-to-SQL with Self-Correction}.
\newblock
\newblock
\showeprint[arxiv]{2304.11015}~[cs.CL]


\end{thebibliography}

\clearpage


\end{document}